\documentclass{article}

\usepackage{graphicx}
\usepackage{listings}
\usepackage{amssymb}

\usepackage{amsmath}
\newcommand{\BigO}[1]{\ensuremath{\operatorname{O}\bigl(#1\bigr)}}

\newtheorem{definicao}{Definition}

\begin{document}

\title{Discovering Semantic Spatial and Spatio-Temporal Outliers from Moving Object Trajectories}

\author{Vitor Cunha Fontes  e Vania Bogorny \\
Universidade Federal de Santa Catarina \\
vitor.fontes@posgrad.ufsc.br ; vania.bogorny@ufsc.br}


\maketitle

\begin{abstract}
Several algorithms have been proposed for discovering patterns from trajectories of moving objects, but only a few have concentrated on outlier detection. Existing approaches, in general, discover spatial outliers, and do not provide any further analysis of the patterns. In this paper we introduce semantic spatial and spatio-temporal outliers and propose a new algorithm for trajectory outlier detection. Semantic outliers are computed between regions of interest, where objects have similar movement intention, and there exist standard paths which connect the regions. We show with experiments on real data that the method finds semantic outliers from trajectory data that are not discovered by similar approaches.

\end{abstract}

\section{Introduction and Motivation}   \label{sec:intro}

The current advances in mobile technology has increased the interest in mobility data analysis in several application domains. Very simple actions as carrying a mobile phone, check emails or to log in social networks software may register the trace of an object. Some devices specially developed for tracking like GPS or sensor networks may capture the movement of people, animals, cars, boats, buses and natural phenomena like hurricanes, storms, etc. The traces collected by such devices are called trajectories of moving objects.

Trajectory data, in general, are raw data, basically containing the identity of the moving object, its position and time. Information as speed, acceleration and direction of the movement can be easily extracted from trajectories.

Several data mining methods have been proposed for discovering different types of trajectory patterns, as groups with similar characteristics \cite{Lee07}, chasing \cite{SiqueiraB11}, flocks  \cite{WachowiczORN11} \cite{LaubeIW05}, sequential patterns \cite{GiannottiNPP07}, periodic patterns \cite{cao2007} \cite{dmkhan2012} and trajectory outliers \cite{LeeHL08} e \cite{Yuan11}. In this paper we are specially interested in trajectory outliers, which are movements that are typically different from the majority of the trajectories in a dataset.

Trajectory outliers can be interesting to discover suspicious behavior in a group of people, to find alternative routes at rush hours in traffic analysis, the best or worst path that connects two districts or regions in a city, fishing vessel or cargo trajectories that scape from their normal route. Outlier discovery can also be interesting in animal movement analysis, to find trajectories that move away from the group, or in soccer analysis, for defining strategies for players to scape from marking.

Existing works for trajectory outlier detection search for patterns in the whole dataset, comparing all trajectories to find those that behave differently from the rest of the group, that are distant from other trajectories or move in different directions. Lee in \cite{LeeHL08} partitions the trajectories in line segments, and outliers are those segments that are distant from the others. The work of  Yuan \cite{Yuan11} differs from \cite{LeeHL08} on the way as trajectories are partitioned. Ge in \cite{GeXZOYL10} divides the space in grids and classifies as outliers the trajectories that are in grids with low density of trajectories. None of the existing approaches consider time, therefore, the discovered outliers are spatial, and no further analysis is performed over the patterns. Time analysis may be very interesting for several applications. In a touristic city, for instance, with high season in January and February, the flow of people at this period will be very different from the rest of the year. The rush hours during the high season may be very different from the normal period, and to find outliers and their meaning can give more possibilities to the decision maker.

While existing approaches search for outliers in the whole trajectory dataset, we address the problem from a different perspective, aiming to discover outliers among trajectories that have the same goal, i.e., trajectories that move between the same regions. In a traffic management application, for instance, one may be interested in finding groups of objects that move together from one region to another in the city, building the most frequent path, and those trajectories that make a different movement between the same regions. In a touristic city, for instance, to find the  most standard path followed by tourists to move around the touristic regions and to discover the trajectories that take alternative and more efficient routes may reveal new points of interest for bus lines, taxi points, emergency routes, etc.

\begin{figure}
\centering
\includegraphics[scale=1.2]{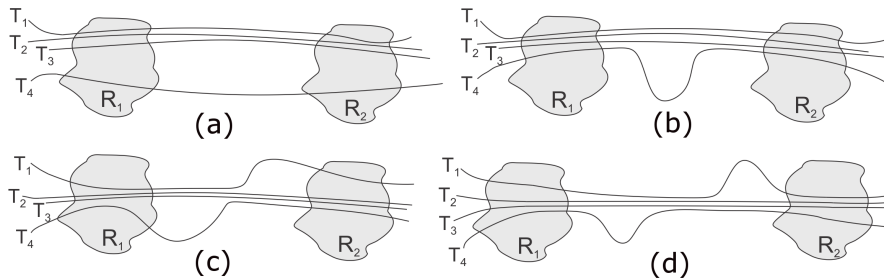}
\caption{Examples of trajectory outliers.}
\label{fig:exPadrao}
\end{figure}

Figure~\ref{fig:exPadrao} shows some examples of outliers. There are four trajectories that move from region $R_1$ to region $R_2$. In Figure~\ref{fig:exPadrao}(a), trajectories $T_1$, $T_2$ and $T_3$ are close to each other, using the same path to move from $R_1$ to $R_2$, characterizing a standard path. Trajectory $T_4$ is far from the group, so we can say that it used a different route, avoiding the rest of the trajectories and becoming an outlier. If we consider that $R_1$ is a Shopping center area in a city and $R_2$ is the downtown region, we may assume that $T_1$, $T_2$ and $T_3$ followed the most frequent route to move between the regions while $T_4$ used an alternative way in its movement. Figure~\ref{fig:exPadrao}(b) shows an example where trajectories $T_1$, $T_2$ and $T_3$ followed the same path, while $T_4$ took an alternative route in the middle of the way from $R_1$ to $R_2$. In Figure~\ref{fig:exPadrao} (c) and (d), the trajectories $T_2$ and $T_3$ followed the same path, while $T_1$ and $T_4$ at some time have chosen a different way.

In this paper we define three novel types of trajectory patterns: standard path, semantic spatial outliers and semantic spatio-temporal outliers. An algorithm named TRA-SOD (Trajectory Semantic Outlier Detection) to find the patterns and to enrich their meaning is presented. In summary, we make the following contributions in relation to existing approaches:
\begin{itemize}
\item Define new types of trajectory patterns.
\item Find both the standard path and the outliers between regions, in both directions.
\item Define a new algorithm for discovering spatial and spatio-temporal semantic outliers.
\item Add meaning to the patterns in an automatic way, analyzing information as duration, traveled distance, stops, and different time granularities.
\end{itemize}

The rest of the paper is organized as follows: Section 2 presents the related works, section 3 presents the main definitions and the algorithm TRA-SOD, while section 4 presents experiments on real trajectory data. Finally, section 5 concludes the paper and suggests directions of future research.

\section{Related Works}

By simply analyzing the physical properties of trajectories it is possible to extract several characteristics about the movement of an object, such as acceleration, speed, displacement, and position. Such information can be used to classify trajectories in pedestrians, cars, ships or planes \cite{DodgeWF09}. This same information can also be used to cluster trajectories \cite{WenLX11}, where those in the same cluster have similar position, velocity and direction.

Some works are concerned only with the position of the trajectory in space and time. In this case, the aim is in finding groups of trajectories that move together, like the moving flock \cite{Laube2004} \cite{WachowiczORN11} and the moving cluster \cite{KalnisMB05}. In the moving flock, the group of trajectories that move together is the same during the entire pattern. In the moving cluster, group members can change since the group keeps a minimum density. These works could be used to find the standard path which connects regions, but apart from not considering regions, the objects in the group must be synchronized during the whole movement, what is not the case in our proposal.

Other methods to group trajectories include \cite{TrasartiPNG11} and \cite{PelekisKKFT11}. In \cite{TrasartiPNG11} the proposal is to extract mobility profiles of moving objects, and for this purpose the method analyzes the trips (or movements) that are close in the space. The similar trips are grouped and the central element of each group, called routine, is used to represent the group. The set of routines of a moving object forms its mobility profile. This work evaluates the trajectories of the same object, while in our case the identity of the object is not important.

The works \cite{Renso2012} and \cite{GiannottiNPP07} consider regions of interest in their analysis. In \cite{GiannottiNPP07} a method is proposed to find sequences of places frequently visited by groups of objects. This method finds the regions of interest considering spatial density or the user can inform these regions. This method groups the trajectories which have similar travel time, i.e., the pattern contains the trajectories which have similar duration to move from one region to another, but the shape of the trajectories that move between the regions (if they move together or not, if they are outliers or not) is not analyzed. Only the travel time is important in this work.

Other papers as \cite{Lee2011} and  \cite{LiLHL09} try to find the paths most frequently used by the moving objects. \cite{Lee2011} maps the trajectories to the streets and analyzes the streets. The sequential patterns are the sequences of roads most frequently used by the moving objects. This approach could also be interesting to find the standard path between regions, but it works only for trajectory data which overlap road networks, and would not be appropriate for trajectories of animals, people in a park, planes, ships and other applications without a road map. Indeed, no regions of interest are considered here. The work of \cite{LiLHL09} was developed for car traffic analysis on road networks. The history of cars moving at different roads is analysed for different timestamps. The cars that behave different from the majority of cars moving on each road at a time are classified as temporal outliers.

The algorithm  proposed by \cite{AlvaresLRB11} finds trajectories that avoid or deviate from target objects as surveillance cameras, traffic jams, or other pre-defined static objects. It verifies for each trajectory if it avoids static objects and if there is a valid path that crosses the region of the avoided object. This work basically differs from our approach because it analyses individual trajectories and does not consider regions of interest.

Knorr in \cite{KnorrN98} defines a method for outlier detection in databases, where outliers are the tuples which are distant from other tuples in the same database. Although this work is not related to trajectories, it was the inspiration for defining the concept of neighborhood in our paper.

Among the works specifically developed for finding outliers in trajectories are \cite{LeeHL08} and \cite{Yuan11}. Both approaches split trajectories in subtrajectories to find outliers. An outlier is a trajectory partition that is far from the majority of the partitions in the dataset. The distance is computed based on the position and the direction of each partition. Outliers are the trajectories where a fraction of partitions are outliers, so it must have a certain length to be an outlier. In both approaches no regions of interest are considered, time is not taken into account and no standard path must exist for trajectories be classified as outliers. The main difference between these approaches is the way as the trajectory partitions (subtrajectories) are generated. None of these algorithms consider any semantic information or give more meaning to the outliers, so we can say that these are geometric outliers, while our work in this paper focuses on semantic outliers.

Some works propose to find outliers in real time trajectories extracted from videos of surveillance cameras, as for instance \cite{GeXZOYL10}, \cite{WangTG06} and \cite{BuCFL09}. In \cite{GeXZOYL10}, the space is divided in grids and the trajectories are analyzed according to the sequence of grids where a trajectory passes by. Two ways are used to find outliers: (i) when the trajectory intersects grids with low density or (ii) when a trajectory follows a direction different from the other trajectories in the same grid. The problem in this work is that only the entire trajectory is analysed instead of partitions, and several patterns may hold in parts of a trajectory. Indeed, the time dimension is not considered by this approach.

The works presented int this section basically identify a common behavior among groups of trajectories or trajectory outliers. None of them look for both spatial and spatio-temporal outliers among trajectories that move between the same regions, and do neither compute nor consider a standard path connecting the regions, as we propose in this paper. Indeed, we compare the traveled distance, the duration and the time granularity of both spatial and spatio-temporal outliers with the standards, in order to infer more semantic outliers and classify them in two types: avoidance outliers and stop outliers.

\section{Mining Semantic Outliers from Trajectories}

This section presents the definitions and an algorithm to find semantic outliers from trajectories.

\subsection{Main Concepts}

Before defining trajectory semantic outliers we present some basic definitions, starting with point.

\begin{definicao}[Point]
A point $p$ is a tuple $(x,y,t)$, where $x$ and $y$ are spatial coordinates and $t$ is the time instant in which the coordinates were collected.
\end{definicao}

A list of points ordered in time is a trajectory.

\begin{definicao}[Trajectory]
A trajectory $T$ is a list of points $\langle p_{1}, p_{2}, p_{3}, ..., p_{n} \rangle$ ,
where $ p_{i} = (x_{i},y_{i},t_{i}) $ and $ t_{1} < t_{2} < t_{3} < ... < t_{n} $.
\end{definicao}

Usually the patterns do not hold for the whole trajectory or during the complete trajectory life. Trajectory patterns occur in part of the trajectories, and this is specially true for outliers. Therefore, we make use of subtrajectories, that is a concept commonly used in trajectory research.

\begin{definicao}[Subtrajectories]
Let $ T = \langle p_{1}, p_{2}, p_{3}, ..., p_{n} \rangle $ be a trajectory. A subtrajectory $S$ of $T$ is
a list of consecutive points $ \langle p_{k}, p_{k+1}, p_{k+2}, ..., p_{m} \rangle $, where $ \forall j,k \leq j \leq m, p_{j} \in T $.
\end{definicao}

\begin{figure}
\centering
\includegraphics[scale=0.3]{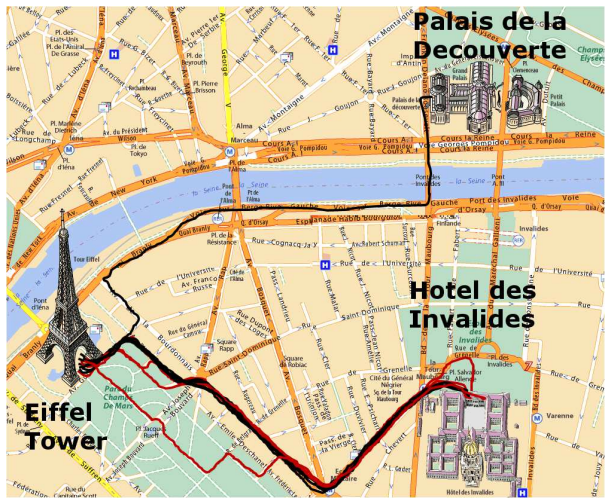}
\caption{Example of candidate.}
\label{fig:exParis}
\end{figure}

In this paper we focus on semantic outliers, and therefore the process starts by looking at subtrajectories that have the same goal, or in other words, that perform a similar movement. By analyzing only subtrajectories that have a common intent of movement increases the certainty of a deviation.  We consider that trajectories that pass by the same regions have a common intent. For instance, objects that move from one district to another in a city, birds that fly from one country to another, or tourists that visit touristic places have the same intent of movement: to go from one region or place to another one. Figure~\ref{fig:exParis} shows an example where a group of trajectories moves from Eiffel Tower to Hotel des Invalides. We can see that these trajectories have the same intent, while the trajectories moving from Eiffel Tower to Palais de la Decouverte have different intent or goal.

To find the movements (subtrajectories) with similar intent we consider regions of interest. These regions can have different size and format, depending on the application. Regions of interest can be districts, dense areas, hot spots, important places, etc. A region can be a pre-defined  important place or computed  by an algorithm that finds dense areas as DBSCAN \cite{Ester96}. How to find these regions is not the focus of this work, but we consider a region as a polygon.

The use of regions allows filtering from the whole dataset only the subtrajectories that move between regions, and outliers will be searched among these sets, what significantly reduces the search space for outliers. We call these subtrajectories that move between regions as candidates. Going back to our example of Figure~\ref{fig:exParis}, the trajectory that moves from Eiffel Tower to Palais de la Decouverte might have done an outlier in relation to the trajectories that move from Eiffel to Hotel des Invalides. However, this  outlier in fact would not be an outlier if we consider that it was the intent of the object to move to Palais de la Decouverte. By considering as start region the Eiffel Tower and the final region the Hotel des Invalides, the candidates will only be subtrajectories that move between these regions, and discarding from the analysis the trajectory going to Palais de la Decouverte. The use of regions of interest apart from being the basis for finding semantic outliers, it significantly reduces the number of trajectories to be analyzed.

\begin{figure}
\centering
\includegraphics[scale=0.2]{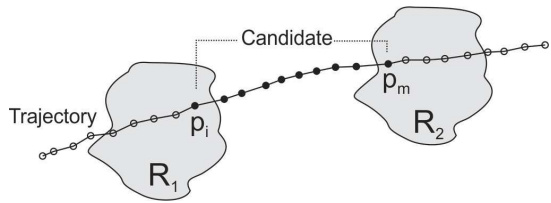}
\caption{Example of candidate.}
\label{fig:candidato}
\end{figure}

We define candidate as the smallest subtrajectory that moves between two regions, i.e., we take the last point of the subtrajectory that intercepts the first region and the first point that intersects the final region, as shown in Figure ~\ref{fig:candidato} .

\begin{definicao}[Candidate]
Let $R_{1}$ and $R_{2}$ be two regions such that $R_{1} \cap R_{2} \neq 0$ and $T$ a trajectory. A candidate from $R_{1}$ to $R_{2}$ is the subtrajectory $S = \langle p_{i}, p_{i+1}, ..., p_{m} \rangle$ of $T$, where $(S \cap R_{1}) = \{p_{i}\}$ and $(S \cap R_{2}) = \{p_{m}\}$.
\end{definicao}

\begin{figure}
\centering
\includegraphics[scale=0.2]{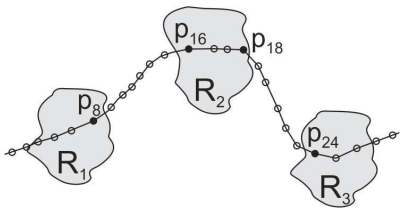}
\caption{Example of a trajectory crossing three regions, generating 3 candidates}
\label{fig:multCandidatos}
\end{figure}

Several regions can be considered for analysis, and it is important to notice that one trajectory may generate several candidates among regions, and one candidate may contain others. An example is shown in Figure~\ref{fig:multCandidatos}, where the same trajectory intersects three regions, and therefore it has three candidate subtrajectories: the subtrajectory that goes from $R_1$ to $R_2$ is a candidate that starts at point $p_8$ and finishes at $p_{16}$. From $R_2$ to $R_3$ is the candidate that starts at $p_{18}$ and finishes at $p_{24}$. From region $R_1$ to $R_3$, $R_2$  is ignored, and a candidate is generated from point $p_8$ to $p_{24}$. In this example, the third candidate contains the two previous ones.

In order to find the outliers, candidates are grouped according to the start and end regions. Each pair of regions generates a different group of candidates. We do not analyze sequences with more than two regions, as for instance the T-pattern \cite{GiannottiNPP07} does, since we are not interested in the outliers between sets of regions.

After defining the set of candidates we start looking for outliers. A candidate will be an outlier when it follows a different path in relation to the majority of the candidates from its group. We can say that a path that is different from the path used by most candidates is of low density, and it has less trajectories in its neighborhood, while a crowded path has many trajectories in its neighborhood. In order to discover these two types of paths we introduce the concept of neighborhood.

For each point of a candidate the neighbors are computed. A candidate is a neighbor of a point if it is close to the point. If a point has a few candidates in its neighborhood, then at that time the moving object was following a path different from the majority of the candidates. The maximal distance for a candidate to be a neighbor of a point is called \emph{maxDist}. In this paper we define neighborhood inspired by the work of \cite{KnorrN98}. The main difference is on the semantics of the neighbor. While in \cite{KnorrN98} the neighbors are points that are close to the object being analyzed, we consider as neighbors the subtrajectories (candidates) that are close to a subtrajectory point.

\begin{definicao}[Neighborhood]
Let $p$ be a point. The neighborhood of $p$ \\* $ N(p,maxDist) = \{ c_i | c_i $ is a candidate and
$\exists q \in c_i , dist(p,q) \leq maxDist \}$.
\end{definicao}

The distance between points $p$ and $q$, defined as $dist(p, q)$, is the euclidean distance between the points.

\begin{figure}
\centering
\includegraphics[scale=0.3]{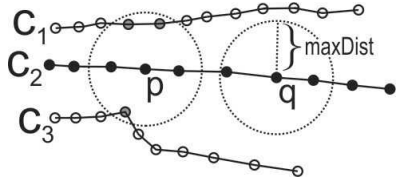}
\caption{Example of neighborhood.}
\label{fig:vizinhanca}
\end{figure}

Figure~\ref{fig:vizinhanca} shows an example of neighborhood.The neighborhood of point $p$ are the candidates $C_1$ and $C_3$, since these two candidates have at least one point inside the radious of size \emph{maxDist} around $p$. Notice that point $q$ has no candidates inside its radios of size \emph{maxDist}, so its neighborhood is empty. We can conclude that at point $p$, $C_2$ was moving with $C_1$ and $C_3$ (same path), but at point $q$, $C_2$ was moving far from $C_1$ and $C_3$ (different path).

To find the path followed by the majority of the candidates, we use the minimum support concept (\emph{minSup}), which is the minimal amount of candidates that a point should have in its neighborhood to be part of a crowded or dense path.  In the example in Figure~\ref{fig:vizinhanca}, if we consider \emph{minSup} 2, the point $p$ in candidate $C_2$ is in a dense path, while point $q$ in $C_2$ moves alone.

The most crowded or dense path between two regions is in general the standard path that connects the regions. A candidate that has all its points in a crowded path is considered as the standard path between two regions. So all candidates that belong to the standard path are called standards.

\begin{definicao}[Standard]
Let $ c = \langle p_{1}, p_{2}, p_{3}, ..., p_{n} \rangle $ be a candidate, $c$ is a standard candidate if and only if $ \forall p_i \in c , \mid N(p_i, maxDist)\mid \geq minSup $.
\end{definicao}

Normally there exists a more frequent path (standard path) to move from one region to another. Going back to our example of Figure~\ref{fig:exParis}, we can see that there is a main path which connects Eiffel Tower and Hotel des Invalides, which is followed by the majority of the trajectories. This path is the standard.

The candidates that have at least one point where cardinality of its neighborhood is less than \emph{minSup} are called \emph{potential outliers}. Therefore, the candidates are split in standards and potencial outliers, so a candidate will always be either a standard or a potential outlier. When all candidates between two regions are  potencial outliers, there is no standard. As a consequence, there is no standard path that an object could avoid or deviate. On the other side, if there is at least one standard, then the potential outliers did really perform a deviation, and are classified as spatial outliers.

\begin{definicao}[Semantic Spatial Outlier]
Let $C$ be the set of candidates between two regions. A potential outlier is a spatial outlier if  $\exists c  \in C | c$ is a standard.
\end{definicao}

The first assumption to define a semantic outlier is that it should move between two regions of interest. The second assumption is that there must be a standard path that connects the regions such that the outlier should avoid or deviate it. Therefore, any subtrajectory that uses a path different from the standard is a semantic spatial outlier.

\begin{figure}
\centering
\includegraphics[scale=0.2]{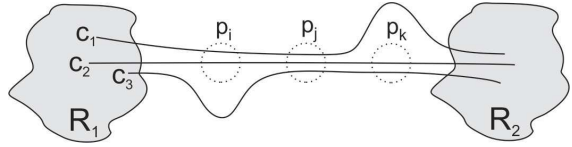}
\caption{Example of Spatial Outlier.}
\label{fig:outlierEspacial}
\end{figure}

Figure~\ref{fig:outlierEspacial} shows an example of semantic spatial outlier, considering $minSup = 1$. In this example there are 3 candidates ($c_1$, $c_2$ and $c_3$) that move from $R_1$ to $R_2$, where, $c_1$ and $c_3$ are spatial outliers and $c_2$ is a standard. This is possible because the neighborhood of $c_2$ is never empty, having at least one candidate. Notice that at $p_j$, both $c_1$ and $c_3$ are in the neighborhood of $c_2$ (the neighborhood is represented by the circle). At $p_i$, $c_1$ is inside the neighborhood of $c_2$, while $c_3$ is doing a deviation. At $p_k$, $c_1$ is deviating, while $c_3$ is moving along $c_2$.

Two candidates leave the same region at the same time window if the difference between the timestamps of the first point of the candidates is less than a given time tolerance. When two candidates leave the start region at the same time window we say that they are synchronized.

\begin{definicao}[Semantic Spatio-temporal Outlier]
Let $C$ be the set of candidates between two regions. A spatial outlier $o$ is a spatio-temporal outlier if $\exists c  \in C | c$ is standard and $c$ is synchronized with $o$.
\end{definicao}

In this work we analyze the time that the objects leave the region, since the objective is to know if they have a syncronized departure, and it is not relevant if they keep the synchronization until reaching the destination. It is quite obvious that outliers which perform a different path will arrive at the destination at a different time than the standards. In this paper we are interested in other time properties of the outliers in relation to the standards, in order to add more semantics to both standards and outliers. We will compare their duration, the traveled distance, the day, month and period that the outlier occurred and if the outlier performed stops during its movement. Based on this analysis, we go one step further to give more semantics to both spatial and spatio-temporal outliers. We classify the outliers in two types: \emph{semantic avoidance outlier} and \emph{semantic stop outlier}. A semantic avoidance outlier is a outlier that has the intent to avoid the group in the standard path to faster reach the destination. Therefore, its duration is lower than the average duration of the standards.

In the next two definitions (Definition~\ref{def:avoidanceOut} and Definition~\ref{def:stopOut}) we will consider that $STA_{O}$  is a set of standards with the same initial and final regions of the outlier $o$. If $o$ is a spatio-temporal outlier, then $STA_{O}$ contains only the synchronized standards, otherwise, if $o$ is a spatial outlier, then $STA_{O}$ contains all the standards. We will also consider that $avgDuration(STA_{O})$ is the average duration of the standards contained in $STA_{O}$.

\begin{definicao}[Avoidance outlier] \label{def:avoidanceOut}
Let $o$ be an outlier, $o$ is a semantic avoidance outlier if $o.duration < avgDuration(STA_{O})$.
\end{definicao}

A semantic stop outlier is a spatial outlier or spatio-temporal outlier that the duration is higher than the average duration of the standards and it made stops in its path. These stops could be either intentional or not, but this issue is out of the scope of this paper. So we can define a stop outlier as:

\begin{definicao}[Stop outlier] \label{def:stopOut}
Let $o$ be an outlier, $o$ is a semantic stop outlier if $o.duration > avgDuration(STA_{O})$ and $o$ has a stop.
\end{definicao}

In this section we presented the basic concepts for semantic outliers, and in the follow section we will show the TRA-SOD algorithm to find this pattern.

\subsection{TRA-SOD: an Algorithm for Mining Trajectory Semantic Outliers}

Listing~\ref{list:algoritmo} presents the pseudo-code of the algorithm TRA-SOD to compute semantic outliers. The input of the algorithm is a set of trajectories $ST$, a set of regions of interest $SR$, the maximal distance (\emph{maxDist}), the minimum support (\emph{minSup}) and \emph{TimeTolerance}. The regions can be places of interest for the application or computed by algorithms as \emph{CB-SMOT} \cite{PalmaBKA08}, \emph{T-patterns} \cite{GiannottiNPP07}, \emph{DB-SCAN} \cite{Ester96}.

The search for outliers between regions has the intent to find objects that cross a region $R_1$ and move to the same destination, a region $R_2$. By doing so, all analyzed trajectories have the same intention, which is to move from the same region to the same destination. This analysis excludes trajectories with different goals and moving far from the interest group of trajectories.
\tiny
\begin{lstlisting}[numbers=left, caption=TRA-SOD, xleftmargin = 22.67 pt, escapeinside={(*@}{@*)},label=list:algoritmo]
INPUT:
ST;  // Set of trajectories
SR;  // Set of regions
maxDist; // maximum distance
minSup;  // minimal number of neighbour
TimeTolerance;

OUTPUT:
Set of semantic spatial and spatio-temporal outliers.

METHOD:
FOR EACH PAR OF REGIONS (startRegion, endRegion) in SR{       (*@\label{line:forSR}@*)
    C = findCandidates(ST, startRegion, endRegion); // find candidates.      (*@\label{line:forcandidato}@*)
    StandardSet = findStandard(C, maxDist, minSup); // find standards. (*@\label{line:forpotential}@*)
    IF ( StandardSet != EmpytSet ) {            (*@\label{line:testeConj}@*)
        SpatialOutSet = C - StandardSet; // Set of spatial outliers   (*@\label{line:spatial}@*)
        FOR EACH SPATIAL OUTLIER out in SpatialOutSet {  (*@\label{line:forspatial}@*)
            out.Time_granularity_refinement;        (*@\label{line:granularity}@*)
            out.Comput_synchronized_standards(TimeTolerance);  (*@\label{line:synchronized}@*)
            IF (out.duration > avg_duration_standards AND out.hasStop) (*@\label{line:stop}@*)
                Out.is("Stop outlier"); (*@\label{line:intentionalstop}@*)
            IF (out.duration < avg_duration_standards)    (*@\label{line:testduration}@*)
                out.is("Avoidance outlier"); (*@\label{line:avoidance}@*)

} } }
return out
\end{lstlisting}
\normalsize

The total number of regions $SR$ is a determinant factor for the processing time of the algorithm, since the trajectories between every pair of regions are analyzed in both directions. For each pair of regions (line~\ref{line:forSR} ), where \emph{startRegion} represents the start region and \emph{endRegion} the final region, the algorithm starts by computing the candidates that move from  \emph{startRegion} to \emph{endRegion} (line~\ref{line:forcandidato}), with the function \emph{findCandidates}. This function checks for every trajectory if it intersects the pair of regions. Once the candidates are computed, the algorithm searches for the standards with the function (\emph{findStandard}) (line~\ref{line:forpotential}), considering the parameters \emph{maxDist} and \emph{minSup}. If the set of standards is not empty (line~\ref{line:testeConj}), then it goes for finding the spatial outliers, since there is a standard path that connects both regions.

Spatial outliers are all candidates which are not standards (line~\ref{line:spatial}). For each spatial outlier (line~\ref{line:forspatial}) the algorithm starts adding semantics to the time dimension (line~\ref{line:granularity}). Instead of simply showing the timestamp of the spatial and spatio-temporal outliers we do interpret the time to make it easier to the user to rapidly identify the periods of the outliers. Therefore, the algorithm extracts from the timestamp in which the standards and outliers live the start region, several information, including: the day of the week (e.g. Monday) that the outlier occurred, the period of the day (e.g. Afternoon), and the month of the year (e.g. November). Such granularity refinement is useful to interpret the patterns.

The next step is to check if the outlier is synchronized with any standard, i.e., if there are standards that leave the start region at a similar time as the spatial outlier (line~\ref{line:synchronized}).  In case there is a synchronized standard, then the spatial outlier becomes a spatio-temporal outlier.

To give more meaning to the outliers the algorithm verifies if the duration is greater than the average duration of the standards (line~\ref{line:stop}). When the outlier is spatio-temporal, the average duration is compared only with the synchronized standards, in order to check if the faster path is the one followed by the outlier or by the standards. If the duration of the outlier is greater, it means that the outlier took more time to move between the regions, and then stops are computed for the outlier (line~\ref{line:stop}). The stops verification can be done by different methods, but in our case we considered the method \emph{CB-SMOT} \cite{PalmaBKA08}.

If an outlier spent more time during his trip and did a stop, it is classified as a Stop Outlier (line~\ref{line:intentionalstop}).

The last step (line~\ref{line:testduration}) is to check if the duration of the outlier is lower than the average duration of the standards, and if this is the case, the outlier is classified as Avoidance Outlier (line~\ref{line:avoidance}), where it had the intent to avoid the group to move faster.

\tiny
\begin{lstlisting}[numbers=left, caption=FindStandard, xleftmargin = 22.67 pt, escapeinside={(*@}{@*)},label=list:findPotencialOutlier]
INPUT:
C  // Set of candidates;
maxDist  // maximun distance;
minSup // minimal neighborhood;

OUTPUT:
Set os standards.

METHOD:
FOR EACH CANDIDATE c in C{       (*@\label{line:eachcandidate}@*)
    FOR ALL POINTS p in c{      (*@\label{line:eachpoint}@*)
        IF( |N(p, maxDist)| > minSup ) (*@\label{line:maxDist}@*)
            ResultSet.add(c); (*@\label{line:result}@*)
    }

}
RETURN ResultSet; (*@\label{line:resultset}@*)
\end{lstlisting}
\normalsize

Listing~\ref{list:findPotencialOutlier} shows the pseudo-code of the function \emph{findStandards}. The input is the set of candidates $C$, the maximal distance \emph{maxDist} for a candidate to be close to another one and \emph{minSup}, that is the minimal number of candidates in the neighborhood for a candidate to be a standard. The output of this function is the set of standards. For each candidate in the set (line~\ref{line:eachcandidate}), and for each point of a candidate (line~\ref{line:eachpoint}) the algorithm checks if the number of neighbors is greater than   \emph{minSup} (line~\ref{line:maxDist}). If this is the case, then the candidate is considered a standard and is added to the \emph{ResultSet} (line~\ref{line:result}).

The complexity of TRA-SOD in the worst case is $\BigO{n^{2}m^{2}}$, where $n$  is the number of points and  $m$ is the number of regions. The algorithm combines all regions (line~\ref{line:forSR}), and for each pair it compares each candidate point (line~\ref{line:maxDist} in function \emph{FindStandard}) with all points in its neighborhood.

\section{Experimental Results}

In this section we present the results of two experiments with real trajectory datasets, compare the results of our method with TRAOD \cite{LeeHL08} and perform a parameter analysis. Both datasets are trajectories of people. The first dataset are trajectories of cars in the city of Porto Alegre, Brazil, and the second represents trajectories of pedestrians in the city of Amsterdam.

\subsection{Experiment 1: Car Trajectories in the City of Porto Alegre}

\begin{figure}
\centering
\includegraphics[scale=0.3]{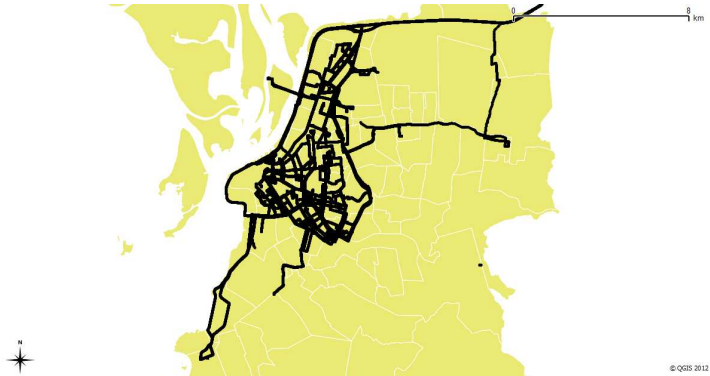}
\caption{Car Trajectories in Porto Alegre.}
\label{fig:trajPortoAlegre}
\end{figure}

This experiment was run on a dataset with 241 trajectories, with a set of 197959 points. The sampling rate of the points is in average every second. Figure ~\ref{fig:trajPortoAlegre} shows this dataset over a map of districts of the city.

This experiment was performed considering 50 meters as the \emph{maxDist}, such that trajectories moving in the same street should be considered as neighbors that follow the same path. Minimum support \emph{minSup} was set to 4, indicating that candidates will be standards if they have at least 4 candidates in its neighbourhood. In this experiment we considered \emph{TimeTolerance} of 30 minutes, since the dataset has not many synchronized trajectories leaving the regions at similar time.

\begin{figure}
\centering
\includegraphics[scale=0.25]{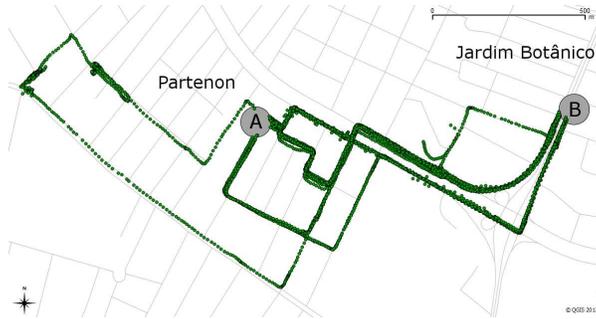}
\caption{Candidates between the regions A and B.}
\label{fig:candidatosPOA}
\end{figure}

The regions considered in this experiment were dense areas in different districts of Porto Alegre. Figure~\ref{fig:candidatosPOA} shows the result between the districts Partenon (called Region A) and Jardim Botanico (called Region B). A total of 24 candidates were found between these regions, among which 8 move from A to B and 16 from B to A. We analyze these candidates separately.

\begin{figure}
\centering
\includegraphics[scale=0.8]{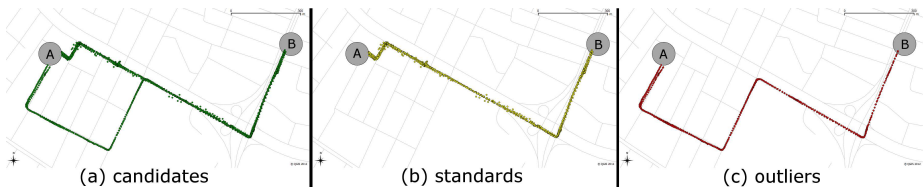}
\caption{(a) candidates moving from A to B; (b) standard path from A to B; (c) outliers moving from A to B.}
\label{fig:compPOA1}
\end{figure}

Figure~\ref{fig:compPOA1}(a)) shows the 8 candidates that move from A to B. In this set, 6 candidates are standards, as shown in Figure~\ref{fig:compPOA1}(b), representing the standard path that connects the two regions, while 2 are outliers, as shown in Figure~\ref{fig:compPOA1}(c).

\begin{figure}
\centering
\includegraphics[scale=0.8]{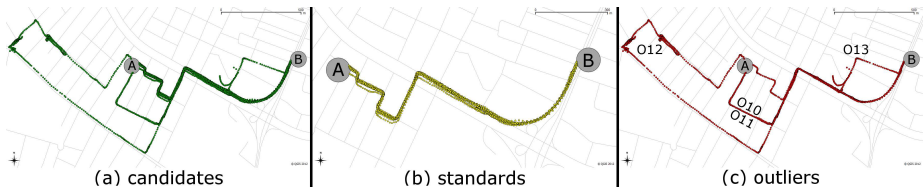}
\caption{(a) candidates moving from B to A; (b) standard path from B to A; (c) outliers moving from B to A.}
\label{fig:compPOA2}
\end{figure}

Figure~\ref{fig:compPOA2} shows the results of the candidates moving in the opposite direction, from region B to A. In this case the algorithm found 16 candidates, shown in Figure~\ref{fig:compPOA2}(a), among which 12 were standards (Figure~\ref{fig:compPOA2}(b)) and 4 were outliers (Figure~\ref{fig:compPOA2}(c)).

Through the figures we can notice that the outliers moving from A to B as well as from B to A follow a longer path than the standards. This is more clear when analyzing the information added to the outliers, as shown in Table~\ref{table:tabelaPOA}, explained in the following section.

\subsubsection{Analyzing the Meaning of Outliers and Standards}

Table~\ref{table:tabelaPOA} presents part of the output of the algorithm TRA-SOD, where the $O3$ and $O4$ are the outliers between the regions A and B, and the others are the outliers between B and A. It presents several characteristics about the outliers and the standards. The first part of the table shows information about the trajectories that are outliers and the second part has information about the trajectories that follow a standard path between the same regions. All outliers (O3, O4, O10, O11, O12, O13) have length (traveled distance) larger than the average length of the standards ( \emph{Avg\_Length}).

The time granularity gives more information about the outliers. Outliers O3 and O4, that move from A to B, occur on Tuesday Morning, in November. The traveled distances (1874,6 and 1883,7 meters) are very similar. Both O3 and O4 are spatio-temporal outliers, having synchronized standards. For each outlier (O3 and O4) there were 2 trajectories following the standard path, leaving Region A at the same time period. The distance traveled by these outliers was around 300 meters longer than the average distance traveled by the standards (1583.69).

\begin{table}
\centering
\includegraphics[scale=3.2]{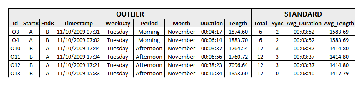}
\caption{Results found by TRA-SOD for $maxDist = 50$, $minSup = 4$ e $TimeTolerance = 30$.}
\label{table:tabelaPOA}
\end{table}

\begin{figure}
\centering
\includegraphics[scale=1.5]{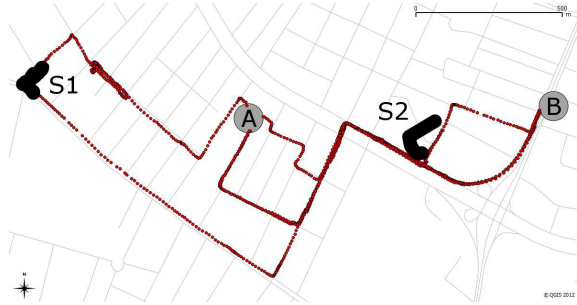}
\caption{Stops of outliers O12 and O13.}
\label{fig:stopPOA}
\end{figure}

The outliers O10, O11, O12 e O13, that move in the opposite direction, from B to A, happened the same day, but at a different period, in the afternoon. From this group of 4 outliers, only O13 is a spatial outlier because no trajectories in the group of standards are synchronized. The others are spatio-temporal outliers. We can notice that outliers O12 and O13 traveled for a much longer period than the standards, taking respectively 38:23 and 16:34 minutes to travel from B to A, while the average duration travel time of the standards was 3:37 and 3:40 minutes. This long trip is explained by the stops of these trajectories. Figure~\ref{fig:stopPOA} shows these stops, where S1 was a stop found at outlier O12 and S2 the stop of outlier O13.  These stops are out of the standard path, and are probably the reason for the outliers. Outliers with a duration significantly higher than the average duration of the standards and that have stops on their subtrajectories, are classified as stop outliers, i.e., outliers that have the intention to stop somewhere else out of the standard path. The stops of outliers O12 and O13 had a duration of 11:22 minutes and 8:34 minutes, respectively, what corresponds to a significant amount of time in the duration of the outlier.

In this experiment we can conclude that, in general, all outliers traveled a longer distance than the standards and took more time in their trips. Among the 6 outliers, two have mad a stop in their alternative route, justifying their outlier path.

\subsubsection{Comparing the algorithms TRAOD and TRA-SOD}

In this section we compare the output of the algorithms TRAOD \cite{LeeHL08} and TRA-SOD. This comparison is performed to show that both methods discover different patterns, which is mainly obvious since the proposals are different. The algorithm TRAOD does not consider regions and it does not perform any further analysis over outliers as TRA-SOD, but in order to compare the results of both algorithms we considered the same trajectory candidates as input for both methods. Different input would generate different output. The TRAOD algorithm has as input the maximal distance between trajectory partitions ($D$), the maximal percentage of trajectories ($p$) for not being outliers and the fraction ($f$) of partitions that a trajectory should have to be an outlier.

\begin{figure}
\centering
\includegraphics[scale=0.8]{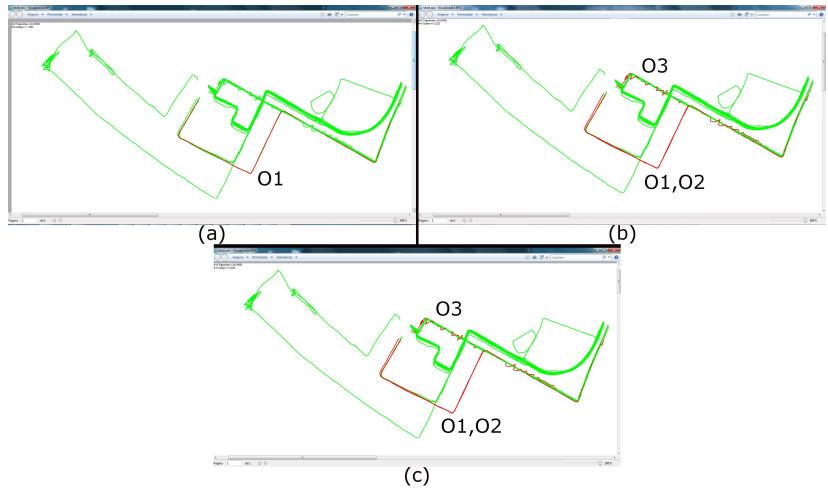}
\caption{Results for TRAOD with parameters (a) D = 82, p = 0.8 and F = 0.2 (b) D = 50, p = 0.8 and F = 0.2 and (c) D = 50, p = 0.8 and F = 0.1.}
\label{fig:traodPOA}
\end{figure}

Figure~\ref{fig:traodPOA} shows the results found with the method TRAOD. The patterns were generated with the algorithm available at \cite{LeeSite}, and we keep the original algorithm output, therefore outliers are shown in red while trajectories are shown green. This algorithm was run with 3 different parameter sets, trying to come as close as possible to TRA-SOD. Figure~\ref{fig:traodPOA}(a) shows the result of TRAOD with the default parameters D = 82, p = 0.8 and F = 0.2, where only one outlier was found. Figures~\ref{fig:traodPOA}(b) and ~\ref{fig:traodPOA}(c) show the results with the distance parameter set as 50 meters, to characterize trajectories in the same street, p = 0.8 for the proportion of trajectories for not being outliers and the fraction of partitions as 0.2 AND 0.1. In both cases only 3 outliers were discovered. It is important to notice that the output of TRAOD is the total number of outliers and the outliers presented over the set of trajectories. No further analysis is performed.

\begin{figure}
\centering
\includegraphics[scale=1.7]{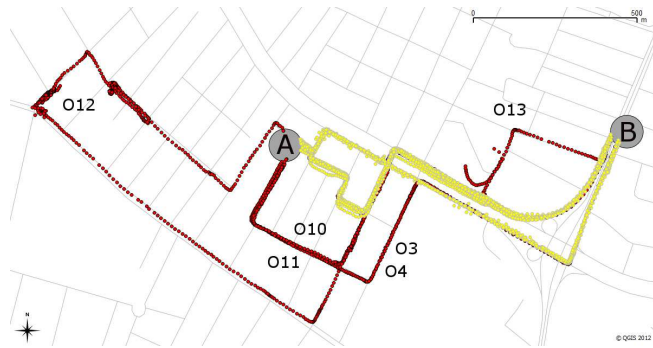}
\caption{Results with the method TRA-SOD.}
\label{fig:resultadoPOA}
\end{figure}

Figure~\ref{fig:resultadoPOA} shows the results of TRA-SOD (where dark trajectories are the outliers and the light ones are the standard path), which found 6 outliers, while TRAOD found at most 3. Indeed, in both cases where TRAOD found 3 outliers, (Figure~\ref{fig:traodPOA}(b) and Figure~\ref{fig:traodPOA}(c)) O3 that was found as an outlier by TRAOD was found by TRA-SOD as part of the standard path.

The difference between the results of both methods can be explained as follows: (i) TRAOD was not developed to discover patterns between regions, (ii) TRAOD does not consider the movement in both ways between two regions, but between all trajectories in the dataset, (iii) no standard path between two regions is considered to find outliers, and (iv) TRAOD generates outliers with a minimal length.

\subsection{Experiment 2: Trajectories of Pedestrians in the City of Amsterdam}

\begin{figure}
\centering
\includegraphics[scale=1.8]{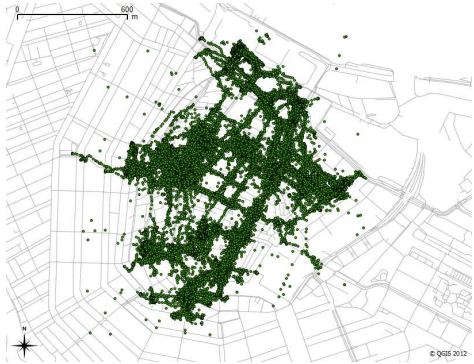}
\caption{Trajectory dataset in the city of Amsterdam.}
\label{fig:trajAmsterdam}
\end{figure}

This experiment was performed with trajectory data of a mobile learning game developed by the Waag Society \cite{Waag} in Amsterdam, with students between 11 and 12 years old. The students were divided in 7 groups, with the objective to
solve a puzzle. Each student got a GPS, which collected his/her trajectory. This dataset is shown in Figure~\ref{fig:trajAmsterdam}.

\begin{figure}
\centering
\includegraphics[scale=1.8]{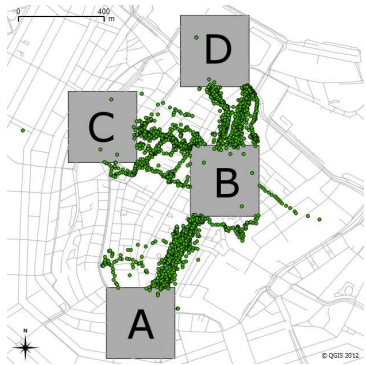}
\caption{Candidates between regions A, B, C and D.}
\label{fig:candidatosAms}
\end{figure}

This dataset has 466 trajectories summarizing 62264 points, with aleatory sampling rates. For this experiment we considered 4 regions with high density. A total of 66 candidates (subtrajectories moving between regions) were computed and are shown in Figure~\ref{fig:candidatosAms}.

The following pairs of regions generated candidates: A to B, B to C, B to D, B to A, C to B and D to B.

TRA-SOD was run considering two different sets of parameters. In both experiments we considered \emph{maxDist} = 50 meters, which is the most appropriate value to group as standards the objects moving in the same street and avoid joining parallel streets. \emph{Timetolerance} was defined as 10 minutes, and \emph{minSup} was considered 4 in one experiment and 6 in the other, as explained in the following sections.

\subsubsection{Case 1}

\begin{figure}
\centering
\includegraphics[scale=1.8]{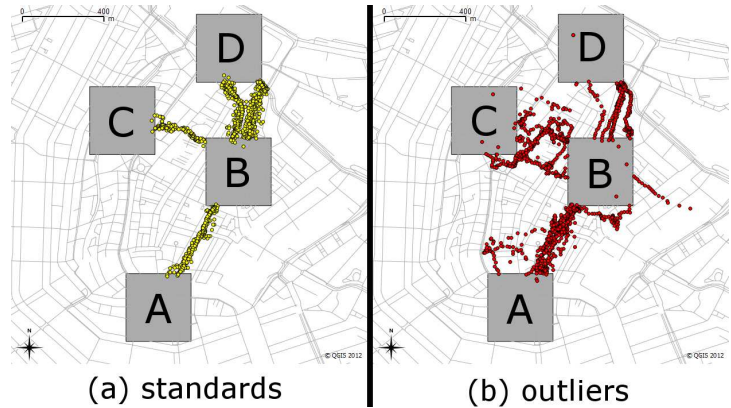}
\caption{(a) Standards and (b) outliers with $minSup = 4$.}
\label{fig:resultadoAms1}
\end{figure}

For $minSup = 4$ a total of 28 outliers and 33 standards were found. Figure~\ref{fig:resultadoAms1} presents the results, where Figure~\ref{fig:resultadoAms1}(a) shows the standards and Figure~\ref{fig:resultadoAms1}(b) the outliers. In Figure~\ref{fig:resultadoAms1}(a) we can notice that the standard path is very clear between the regions. It is interesting to see that there is one standard path that connects A to B and B to C, while regions B and D have 2 standard paths. In Figure~\ref{fig:resultadoAms1}(b) we can see that the outliers are sparse in the city, using alternative ways to move between regions.

\begin{figure}
\centering
\includegraphics[scale=1]{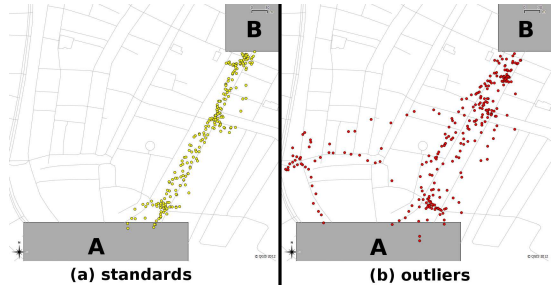}
\caption{(a) Standards and (b) outliers moving from region B to A.}
\label{fig:exAms1}
\end{figure}

Table~\ref{table:tabelaResultadoAms1} shows the distribution of outliers and standards between the regions. For the 10 candidates between B and A, 5 are standards and 5 are outliers, i.e., 5 use a standard path while other 5 trajectories choose alternative routes, as can be seen by Figure~\ref{fig:exAms1}. In Figure~\ref{fig:exAms1}(b) we can notice that the outliers follow a sparse path in horizontal directions in relation to the standards (Figure~\ref{fig:exAms1}(a)).

\begin{table}
\centering
\includegraphics[scale=4]{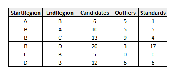}
\caption{Outliers and standards per region.}
\label{table:tabelaResultadoAms1}
\end{table}

\begin{figure}
\centering
\includegraphics[scale=1]{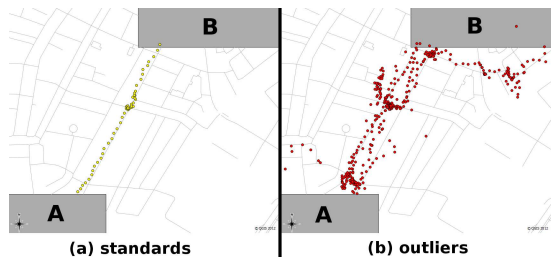}
\caption{(a) Standards and (b) outliers moving from region A to B.}
\label{fig:exAms2}
\end{figure}

Figure~\ref{fig:exAms2} shows a case between regions A and B where there is only one standard (Figure~\ref{fig:exAms2}(a)), but 5 outliers (Figure~\ref{fig:exAms2}(b)). This occurs because for all trajectory points of the standard, the number of neighbors is higher than \emph{minSup}, i.e., the standard is always moving together with at least 4 candidates. This is a case similar to the one shown in Figure~\ref{fig:outlierEspacial}. Although with $minSup = 4$ there is only one trajectory in the standards, this happens because all neighbors become outliers, having deviated the main route at some time during their movements. This is understandable because every outlier follows a different route at a moment.

From B to D, 17 subtrajectories follow the standard path and 3 are outliers.

Still in Table~\ref{table:tabelaResultadoAms1} we can notice that among the 5 candidates that move from C to B no standard path was found, as a consequence, no outlier will emerge.

Table~\ref{table:tabelaFinalAms1} shows the standards and outliers found in this experiment. The average duration and length of the outliers can be compared with the synchronized standards, which are shown in gray color. When the number of synchronized standards is zero, then the attribute values refer to all standards. For instance, Outlier O9 has no synchronized standard, therefore the average duration and length of the standards  refers to all 6 standards that move between the same regions as O9. Among the 28 outliers, five (O7, O8, O28, O30 and O51) are spatio-temporal because they have synchronized standards. The outliers happened in two different days (Thursday and Friday), in June, at different periods of the day (morning, evening and night). No outlier was found in the afternoon, and all outliers on Fridays were at morning or night, while outliers on Thursday  happened at night.

\begin{table}
\centering
\includegraphics[scale=3.2]{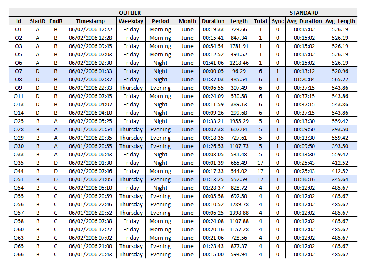}
\caption{Outliers for $minSup = 4$.}
\label{table:tabelaFinalAms1}
\end{table}

Among the 28 outliers, 14 (O5, O7, O9, O11, O13, O14, O28, O33, O35, O44, O51, O55, O56 and O57) were faster than the average duration of the standards, and are therefore classified as avoidance outliers, for having avoided the standard path. Among these 14 outliers, 7 (O5, O7, O9, O11, O13, O14 and O33) traveled a shorter distance than the average distance of the standards, what shows that these students followed a shorter path to reach the destination. The other 14 outliers (O1, O2, O3, O6, O8, O25, O29, O30, O54, O58, O60, O63, O65 and O66) were slower in their movements, and five of them  (O3, O6, O8, O54 and O65) had a stop on their subtrajectories, and therefore are classified as stop outliers.

\subsubsection{Case 2}

\begin{figure}
\centering
\includegraphics[scale=1.4]{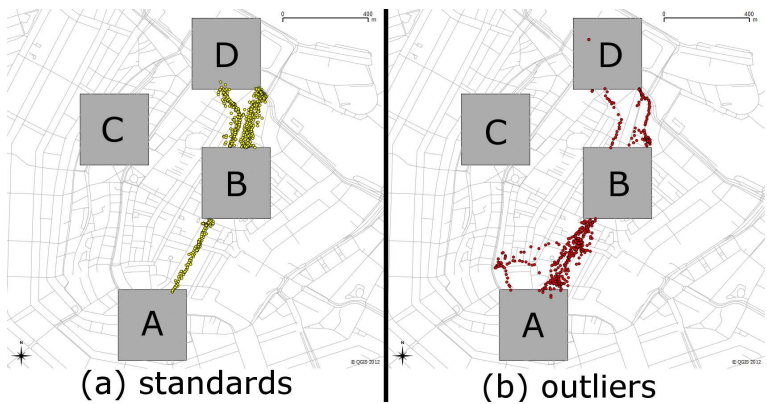}
\caption{(a) Standards and (b) outliers with $minSup = 6$.}
\label{fig:resultadoAms2}
\end{figure}

A second experiment considering $minSup = 6$ found a total of 19 standards and 11 outliers, as shown in Table~\ref{table:tabelaResultadoAms2} and Figure~\ref{fig:resultadoAms2}. With an increase of \emph{minSup} more candidates are needed to find the standard path, what of course resulted in the reduction of the number of standards.  Among the 5 standard paths found in Case 1, only 2 were found in case 2: from B to A and from B to D. Table~\ref{table:tabelaResultadoAms2} shows the total of standards and outliers between these regions. From B to A 3 standards and 7 outliers were discovered. This is possible because among the 10 candidates, only 3 standards moved through the most direct path, as shown in  Figure~\ref{fig:resultadoAms2}.  From B to D a set with 16 standards was found and 4 outliers, what means that most candidates moved together over the same path. From B to A and from B to D the number of outliers increased when compared to the result of Case 1, with lower minimum support.

\begin{table}
\centering
\includegraphics[scale=4.5]{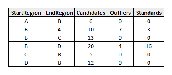}
\caption{Outliers and standards per region.}
\label{table:tabelaResultadoAms2}
\end{table}

Table~\ref{table:tabelaFinalAms2} shows the details about the outliers from regions B to A (O24, O25, O28, O29, O30, O31 and O33) and from  B to D (O35, O37, O44 and O51), where the rows in gray highlight spatio-temporal outliers. We can notice that most outliers found for minimal support 6 were also found with support 4, what shows that  basically the \emph{minSup} affects the discovery of the standard path, and not the outliers. Only 3 outliers (O24, O31 and O37) were not found in Case 1.

Among the outliers in Table~\ref{table:tabelaFinalAms2}, five were on Thursday night and six on Friday morning and night. Among the outliers on Thursday (O24, O28, O29, O30 and O51), their path was longer than the path followed by the standards and the duration was similar, except for outlier O30 that had a duration of 1:26 hours, while the duration of the standards was 9:50 minutes. All outliers on Friday are spatial only, since there is no synchronized standard moving at similar time.

Seven of the 11 outliers (O28, O31, O33, O35, O37, O44 and O51) are classified as avoidance outliers, since they moved faster than the standards. Among the slower outliers non of them have made stops, what means that they selected a worse path to follow in terms of speed of movement.

\begin{table}
\centering
\includegraphics[scale=3.2]{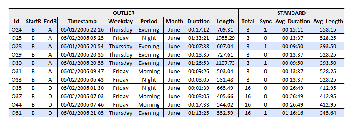}
\caption{Outliers for $minSup = 6$.}
\label{table:tabelaFinalAms2}
\end{table}

\subsubsection{Comparing TRAOD and TRA-SOD}

\begin{figure}
\centering
\includegraphics[scale=1.5]{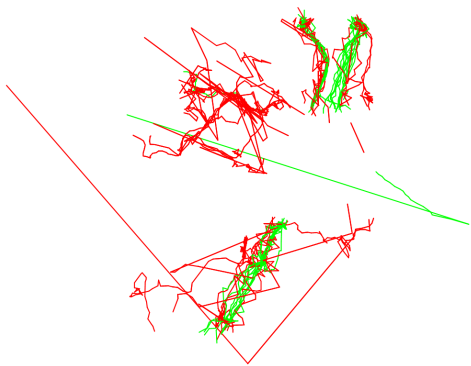}
\caption{TRAOD (F = 0.2; D = 82; p = 0.95).}
\label{fig:traodAms}
\end{figure}

Several experiments were performed with TRAOD and the closest result to TRA-SOD  is shown in  Figure~\ref{fig:traodAms}. As the experiments with TRAOD were done with the algorithm of the authors (downloaded from \cite{LeeSite}), we did not edit the resulted image to add the regions to the figure, since TRAOD does not consider regions. TRAOD transforms subtrajectories in lines, what makes the result a bit different from TRA-SOD. TRAOD found a total of 36 outliers among the candidates, while TRA-SOD found 28 and 14. TRAOD does not look for a standard path to find outliers, and  does not do any further analysis, therefore it is impossible to infer the direction of the outliers found by TRAOD, if they are going or returning, and if they are synchronized or  happened at a similar period,  since time is not considered.

\subsection{TRA-SOD Parameter Analysis}

As in any data mining algorithm, the parameter definition is a concern, and they directly affect the result of the algorithm. TRA-SOD makes use of three parameters only: \emph{maxDist}; \emph{minSup}; \emph{TimeTolerance}.  \emph{maxDist} is used to check if trajectories use the same path to find the standard route between two regions. The best value  for this parameter is the width of the path used by the trajectories. In case objects are moving on road networks, \emph{maxDist} should be as large as the street, such that only trajectories on the same street will be grouped in the same path. In applications where objects do freely move as birds in the sky or animals in the woods or even pedestrians in a park, \emph{maxDist} may vary and can be set according to what the user defines as the width of a path. A to small \emph{maxDist} may split objects that move in the same path, making it more difficult to find the standards. A very high \emph{maxDist} may join objects that move in different paths (distant paths) in the same group. Therefore, this parameter depends on the application.

Table~\ref{table:tabelaAnalise1} presents an analysis of the parameters for the Amsterdam dataset, for $minSup = 4$, $TimeTolerance = 10$ minutes and varying \emph{maxDist}. Three different values were considered for \emph{maxDist}: 20, 50 e 80 metros. The first two columns are the start and end region. Only paths which had standards are shown in the table. The number of candidates has not changed, since it does not depend on minimum distance. As can be noticed, the smaller the \emph{maxDist}, the more difficult to find the standard path, since trajectories must be closer to be considered as moving in the same path. Therefore, when  $maxDist = 20$, only two paths (from B to A and B to D) have at least on standard, while five paths have at least one standard for $maxDist = 50$ and $80$. By looking only at the paths which have standards we may notice that the higher \emph{maxDist}, the more candidates follow the same path, increasing the number of standards and reducing the outliers. So the smaller the minimal distance, the lower the number of standards and more outliers will be found, and vice-versa.

\begin{table}
\centering
\includegraphics[scale=4]{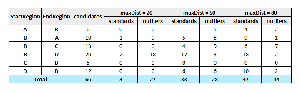}
\caption{Results for TRA-SOD with  $minSup = 4$ and $TimeTolerance = 10$ minutes.}
\label{table:tabelaAnalise1}
\end{table}

Table~\ref{table:tabelaAnalise2} presents the results for $maxDist = 50$ meters, $TimeTolerance = 10$ minutes and varying \emph{minSup}. We considered \emph{minSup}: 2, 4 and 6. Table~\ref{table:tabelaAnalise2} shows that for a low \emph{minSup} (2) more candidates (A to B, B to A, B to C, B to D and D to B) generate a standard path, in total 44 standards. Increasing \emph{minSup} (6), decreases the number of standards (B to A and B to D) to 19. The lower the \emph{minSup} the higher the number of standards and less outliers. The higher the \emph{minSup} the lower the number of standards and more outliers. In the last case, the number of standards may reduce to zero, what means that no standard path was found. As a consequence, no outliers are generated (see rows 1, 3, 5 and 6 in table~\ref{table:tabelaAnalise2}).

The minimal support is also application dependent, so it can be high for a dataset where dense regions have several trajectories.

\begin{table}
\centering
\includegraphics[scale=4]{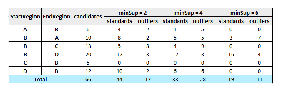}
\caption{Results for TRA-SOD with $maxDist = 50$ e $TimeTolerance = 10$ minutes.}
\label{table:tabelaAnalise2}
\end{table}

The \emph{TimeTolerance} influences the amount of spatio-temporal outliers. The higher the \emph{TimeTolerance} the higher is the chance for several trajectories being traveling within the time window. However, a very high \emph{TimeTolerance} may be meaning less in the sense that trajectories should be moving together.

\subsection{Conclusion and Future Works}

Several algorithms have been proposed for trajectory data mining, but only a few consider trajectory outlier detection and pattern interpretation.

Existing approaches for trajectory outliers mainly consider the space dimension, and time has not been the focus. They do not make deeper analysis of the discovered patterns to give more
meaning or semantics. Indeed, existing trajectory outlier algorithms search for detours in the whole dataset. In this paper we look for outliers among trajectories that have similar goals, i.e., trajectories that move between the same regions. We present de definition of semantic outlier and an algorithm to find such patterns from trajectories, considering both spatial and spatio-temporal information. Semantic outliers are those trajectories that move between regions of interest and that follow a path different from the standard route that connects the regions. The proposed algorithm, named TRA-SOD, finds both the standard path which connects the regions and the outliers.

For all outliers the proposed method evaluates the duration, the traveled distance, the time of the outliers in relation to the standard path, and if outliers made stops, giving more semantics to both spatial and spatio-temporal outliers. With such information we classify the patterns in avoidance outliers and stop outliers.

In this paper we first look for spatial outliers and in a second step verify if they are synchronous in time, i.e., if they leave the region of interest at the same time window as the standards. We do so because only if patterns are spatial they may be temporal or not. By looking only to spatio-temporal outliers several patterns that occur in space but do not hold for similar time period would be missed.

We compared the proposed method TRA-SOD with the algorithm TRAOD to show that the results are different, since both methods have been developed for different goals. Although TRA-SOD looks for patterns among trajectories which travel between the same regions instead of the whole dataset, the running time of TRA-SOD is higher than TRAOD, since it has to find the candidates and to interpret the outliers.

Future ongoing work includes a deep analysis on the standards and the use of clustering techniques to group trajectories that are in the same path. In addition, we will define more types of outliers and perform more experiments on generated data, where the outliers are previously known. We will also do further analysis on the interpretation of the outliers, giving weights for avoidance outliers and stop outliers.

\end{document}